\documentclass[conference]{IEEEtran}
\IEEEoverridecommandlockouts
\usepackage{cite}
\usepackage{amsmath,amssymb,amsfonts}
\usepackage{graphicx}
\usepackage{textcomp}
\usepackage{xcolor}
\usepackage{multirow}
\usepackage{listings}
\usepackage[normalem]{ulem}
\usepackage{hyperref}

\usepackage{xcolor}

\definecolor{codegreen}{rgb}{0,0.6,0}
\definecolor{codegray}{rgb}{0.5,0.5,0.5}
\definecolor{codepurple}{rgb}{0.58,0,0.82}
\definecolor{backcolour}{rgb}{0.95,0.95,0.92}
\lstdefinestyle{mystyle}{
    backgroundcolor=\color{backcolour},   
    commentstyle=\color{codegreen},
    keywordstyle=\color{magenta},
    numberstyle=\tiny\color{codegray},
    stringstyle=\color{codepurple},
    basicstyle=\fontsize{7}{7}\selectfont\ttfamily,
    breakatwhitespace=false,         
    breaklines=true,                 
    captionpos=b,                    
    keepspaces=true,                 
    numbers=left,                    
    numbersep=5pt,                  
    showspaces=false,                
    showstringspaces=false,
    showtabs=false,                  
    tabsize=2
}

\hypersetup{
    colorlinks=true,
    linkcolor=red,
    anchorcolor=black,
    citecolor=green,
    menucolor=red,
    runcolor=magenta,
    filecolor=magenta,      
    urlcolor=blue,
    pdftitle={IS-WIN Research},
    pdfpagemode=FullScreen,
    }

\def\BibTeX{{\rm B\kern-.05em{\sc i\kern-.025em b}\kern-.08em
    T\kern-.1667em\lower.7ex\hbox{E}\kern-.125emX}}

\renewcommand{\baselinestretch}{0.94}
    
\begin{document}

\title{Hardware Acceleration for Real-Time Wildfire Detection Onboard Drone Networks
\thanks{This material is based upon work supported by the Air Force Office of Scientific Research under award number FA9550-20-1-0090, the National Aeronautics and Space Administration (NASA) under award number 80NSSC23K1393, and the National Science Foundation under Grant Numbers CNS-2232048, and CNS-2204445.}
}

\author{\IEEEauthorblockN{Austin Alexander Briley, Fatemeh Afghah}
\IEEEauthorblockA{Holcombe Department of Electrical and Computer Engineering, Clemson University, Clemson, SC, USA \\
\{aabrile,fafghah\}@clemson.edu}
}

\maketitle

\begin{abstract}

Early wildfire detection in remote and forest areas is crucial for minimizing devastation and preserving ecosystems. Autonomous drones offer agile access to remote, challenging terrains, equipped with advanced imaging technology that delivers both high-temporal and detailed spatial resolution, making them valuable assets in the early detection and monitoring of wildfires. However, the limited computation and battery resources of Unmanned Aerial Vehicles (UAVs) pose significant challenges in implementing robust and efficient image classification models. Current works in this domain often operate offline, emphasizing the need for solutions that can perform inference in real time, given the constraints of UAVs. To address these challenges, this paper aims to develop a real-time image classification and fire segmentation model. It presents a comprehensive investigation into hardware acceleration using the Jetson Nano P3450 and the implications of TensorRT, NVIDIA's high-performance deep-learning inference library, on fire classification accuracy and speed. The study includes implementations of Quantization Aware Training (QAT), Automatic Mixed Precision (AMP), and post-training mechanisms, comparing them against the latest baselines for fire segmentation and classification. All experiments utilize the \textbf{FLAME} dataset - an image dataset collected by low-altitude drones during a prescribed forest fire, focusing on key performance metrics such as latency, Mean Pixel Accuracy (MPA), Mean Intersection over Union (MIOU), Frames Per Second (FPS), batch size, throughput, and memory utilization (Active Memory, Allocator State). This work contributes to the ongoing efforts to enable real-time, on-board wildfire detection capabilities for UAVs, addressing speed and the computational and energy constraints of these crucial monitoring systems. The results show a 13\% increase in classification speed compared to similar models without hardware optimization. Comparatively, loss and accuracy are within 1.225\% of original values. The provided source code and additional information are available on the IS-WIN Fire Classification Research page. \footnote{\url{https://github.com/Austin-TheTrueShinobi/IS-WiN-Research}}.

\end{abstract}


\section{\textbf{Introduction}}

Wildfire devastation continues to escalate. Traditional methods reliant on satellites often suffer from significant delays in detecting fires, particularly in remote and forested areas. Autonomous drones equipped with advanced sensors emerge as a promising solution for early fire detection, offering unparalleled high temporal and spatial resolution imaging. Recent works \cite{9217742,afghah2019wildfire,shamsoshoara2021aerial} have explored the potential of deep learning-based fire detection using aerial images collected by drones. Recognizing the absence of wide-bandwidth communication in remote areas,  the potential for immediate communication between drones and fire management centers is constrained. Therefore, processing the collected videos or images onboard before sending them to ground stations is critical, necessitating power-efficient and capable GPUs such as the Jetson Nano. However, real-time models face various hurdles including limitations in onboard processing power, and battery life.  While prior works exploring onboard drone processing exist \cite{afghah2019wildfire}, most have not delved into the realm of real-time fire detection and classification.

Several studies have tackled deep learning acceleration in embedded systems: quantization, pruning, and hardware co-processing being prominent examples \cite{9103821,10318777}. This research, however, identifies a gap in focusing on hardware acceleration for efficient and accurate fire detection onboard UAVs, prioritizing speed. Here, we explore the potential of activation functions (ELU, ReLu, PReLu) and their impact on classification training and memory efficiency utilizing a UAV-collected forest fire dataset called \textit{FLAME} \cite{shamsoshoara2021aerial}. Analyzing them against both hardware and software-accelerated operations is crucial for selecting the optimal combination for fire classification.
We should note that while alternative acceleration techniques like pruning exist, activation functions offer a versatile and straightforward approach with minimal impact on model architecture \cite{9461312}. Each function possesses unique characteristics: ReLU's efficiency can be hampered by dead neurons, ELU's smoothness tackles complex patterns, and PReLU's learnable parameter addresses dead neurons.

We also developed a custom CUDA kernel to parallelize intensive operations on the Jetson Nano's tensor cores. Inference optimizations and post-training quantization (specifically FP16) were chosen for their hardware support and robustness within the training loop.
All configurations are evaluated and compared against multiple architectures, with a special focus on a recent fire-segmentation model created by merging \textit{MobilenetV3} and \textit{DeepLabV3+}\footnote{ \url{https://github.com/maidacundo/real-time-fire-segmentation-deep-learning/tree/main}.} \cite{LI2022145}. Additionally, the performance achieved on the Jetson Nano is compared to prior work on higher-end GPUs, such as the fire-segmentation methods 2080-ti, with limited training image memory to that obtained using a desktop GPU with access to the entire dataset. This provides valuable insights into the trade-offs involved in deploying fire detection models on resource-constrained platforms.
Extensive experimental results show that the proposed model holds significant promise for advancing real-time, onboard fire detection with drones, empowering quicker wildfire management responses via faster inference by incorporating TensorRT. 
Proven Past TensorRT optimization results illustrate optimized compression and FPS improvements of nearly 40\% \cite{10318777}. The integration of TensorFlow-TensorRT (TF-TRT) for low-latency inference has emerged as a key optimization strategy \cite{10318777}.

\section{Related Work}
The need for accurate and real-time fire detection has driven advancements in diverse domains, including deep learning and resource-constrained platforms. This work aligns with several key threads of research:

\subsection{Offline Deep Learning for Fire Detection using Aerial Images}
 CNN-based frameworks have demonstrated promising results in detecting early forest fires \cite{boroujeni2024comprehensive}. These studies validate the feasibility of deep learning for fire detection and provide potential avenues for model adaptation or joint dataset initiatives. Existing research emphasizes the advantages of drones for early fire assessment in remote areas. Furthermore, the presented multi-modal UAV dataset with RGB and thermal images offers a valuable resource for future research and potentially for validating or adapting fire detection models \cite{Boone,chen2022wildland,LI2022145}.

\subsection{Acceleration Strategies for Real-Time Image Processing}
Several recent works have been developed addressing inference acceleration on FPGAs \cite{10.1145/3289185}. While focusing on human activity classification with radar data, the work on hardware acceleration for CNNs on FPGAs shares similarities with this research focus on real-time tasks with resource constraints \cite{8753553}. Their findings regarding parallel processing, data quantization, and decision optimization highlight the potential of hardware acceleration for efficient real-time applications \cite{10031597,10318777}.

\subsection{Contributions of the proposed work}
This paper proposes a real-time fire classification and segmentation model by exploring activation function optimization and NVIDIA Open-source SDKs to accelerate fire classification speed on the Jetson Nano, a resource-constrained platform suitable for drone deployment. The creation of a custom CUDA kernel driver that maps and optimizes for classification speed, lower power consumption, and memory management callback reductions are of primary contributions. This work adds to related works via the perspective of inclusion with quantization techniques - selective quantization, the impact of various activation functions with a curated AMP and post-training quantization (PTQ) function block for a UAV-collected image dataset-FLAME, and the evaluation of these optimizations tailored for both classification and inference tasks in flame classification. It contributes to the field of fire detection by exploring optimization strategies tailored for low-power embedded systems while addressing the critical need for real-time performance. This specific optimization approach, analyzing the interplay of ELU, ReLu, and PReLu with memory efficiency and training, is not addressed in the presented related works.

\section{Intelligent Fire Detection and Analysis: A Combined Classification and Segmentation Model}
\begin{figure*}
    \begin{center}
      \includegraphics[scale = .32]{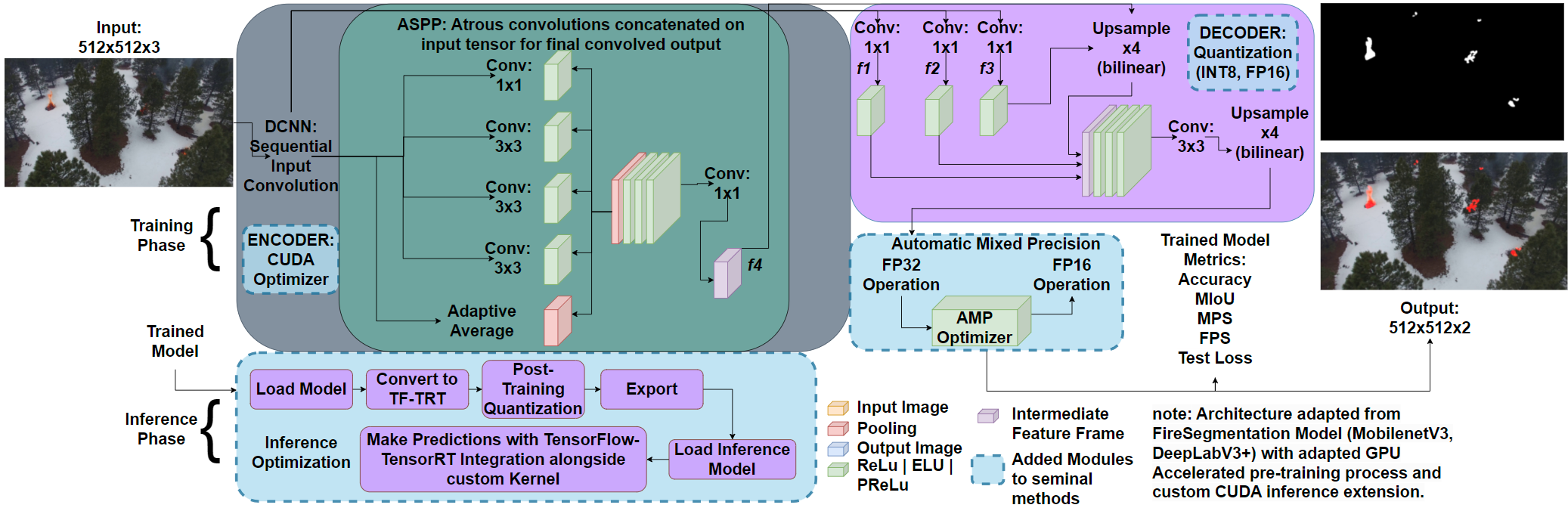}
      \caption{{\small An overview of the classification and segmentation framework used for training and inference. The model is adapted from the amalgamated \textit{MobilenetV3} and \textit{DeepLabV3+} architecture used in \cite{LI2022145}. Modifications to the training loop incorporate AMP and Quantization Aware Training for emulating inference time. The induced error from both AMP and quantization, during post-training, are mitigated by this modeling, allowing it to mitigate the error. The trained model is then converted for TensorFlow-TensorRT (TF-TRT) inference on the NVIDIA TAO toolkit.}}
      \label{fig:mainframework}
  \end{center}
  \vspace{-10pt}
\end{figure*}
The overall model proposition is to further optimize wildfire detection speed without reducing accuracy. The architecture of the fire classification and segmentation model developed in \cite{LI2022145}, which serves as the foundational fire segmentation model for our study along with the proposed modification, is illustrated in Fig. \ref{fig:mainframework}.

\subsection{Model Architecture Optimization and Training Procedure Overview} For semantic segmentation, the DeepLabV3+ model modification consists of an encoder and a decoder. The encoder comprises a Deep Convolutional Neural Networks (DCNN) backbone and an Atrous Spatial Pyramid Pooling (ASPP) module, while the decoder restores features to the original image size. The dataset is split into 85\% training, 15\% validation, and 15\% testing, with shuffling based on the original seed. Data augmentation, including perspective distortion and random transformations, is applied during training. 

\textbf{Encoder}. The DCNN backbone features a standard convolutional layer with 16 convolution filters and 15 MobileNetV3 bottlenecks, generating three intermediate feature maps. ASPP includes a 1x1 convolution, three 3x3 atrous convolutions with varying dilation rates, and an image pooling layer \cite{9598928}.

\textbf{Decoder}. A 1x1 convolution adjusts the channel numbers of features. Features are upsampled to match the size of intermediate features. Concatenation and a 3x3 convolution adjust channels to 2 for background and foreground masks. Final spatial features are upsampled to the original image size. The training uses a batch size of 2, with train, validation, and test data loaders created from the split dataset. The baseline approach employs the Lion optimizer with specific learning rate schedulers and a checkpoint monitor. The number of epochs is set to 30, with validation every 200 batch steps and at the end of each epoch.

\subsection{Hardware Testbed} The hardware of choice, the NVIDIA Jetson Nano paired with CUDA 12.3, excels for its suitability in drone-based fire detection. Its compact size minimizes payload weight, maximizing flight time and reducing energy consumption. This is crucial for drones patrolling vast and often remote areas. Furthermore, the Jetson Nano's energy-efficient architecture balances processing power with low power draw, ensuring extended battery life on resource-constrained platforms. Onboard memory enables real-time image classification inference directly on the drone, eliminating dependence on high-bandwidth communication and ground station infrastructure, thus facilitating faster response times and improved situational awareness. CUDA compatibility simplifies programming for the Jetson Nano's GPU, allowing us to leverage its parallel processing capabilities for efficient computation of convolutional operations within our deep learning models. In essence, the Jetson Nano provides a balance of power, efficiency, and portability, making it an ideal platform for real-time, onboard wildfire detection with drones.

\section{Boosting UAV-Based Fire Detection Speed: A Multi-Pronged Approach with TensorRT and Quantization}

\begin{figure}[htbp]
    \begin{center}
      \includegraphics[scale = .15]{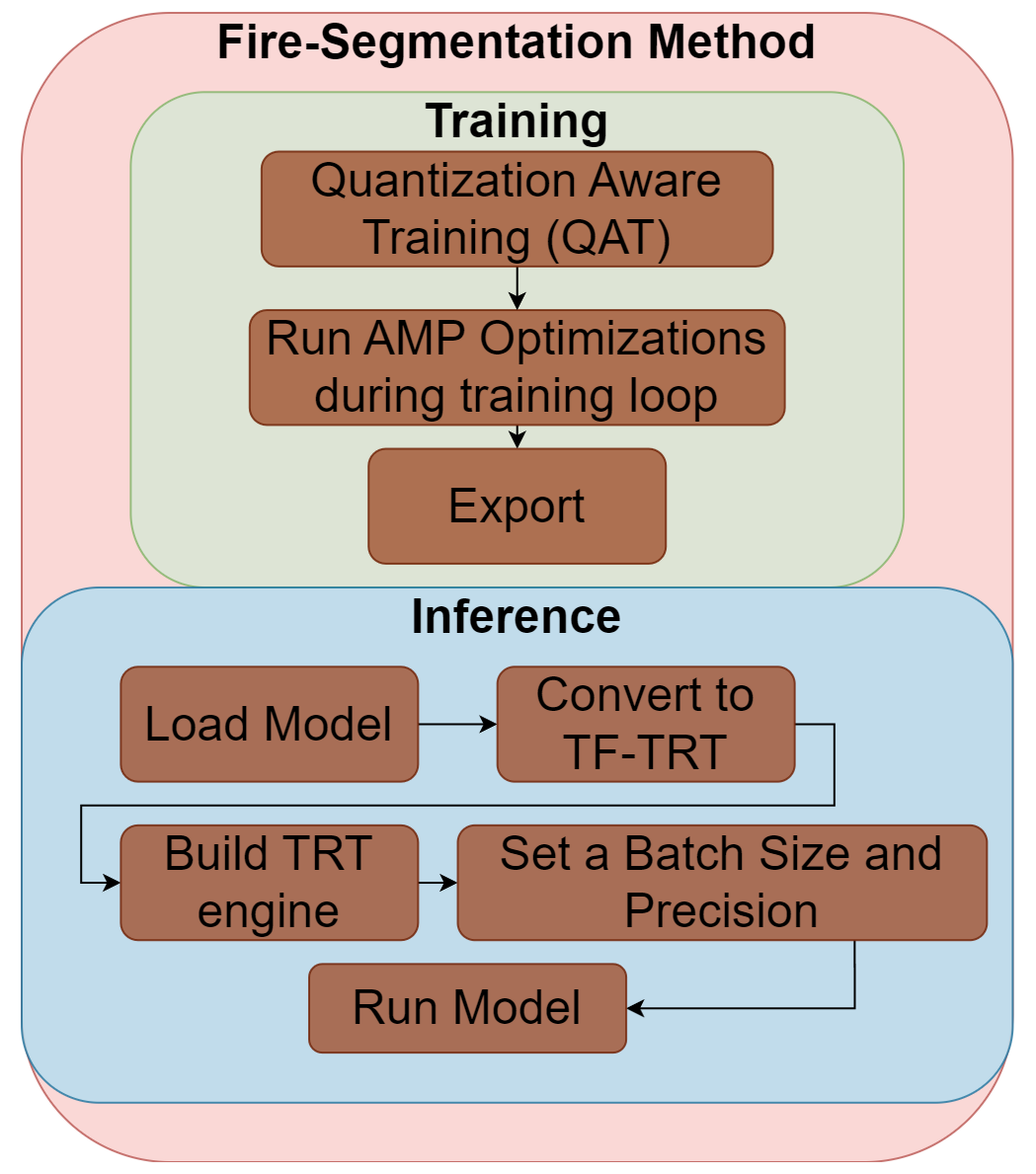}
      \caption{{\small The proposed methods for fire-segmentation inclusion consist of both training and post-training adjustments. Utilizing QAT and AMP ensures proper model preparation for quantization and model inference speed.}}
      \label{fig:methodprop}
  \end{center}
  \vspace{-10pt}
\end{figure}

In this section, we discuss a comprehensive set of proposed optimization methods specifically tailored for wildfire detection on the NVIDIA Jetson Nano illustrated in Figure \ref{fig:methodprop} to address the challenges of resource-constrained drone network efficiency.
This approach is centered around several critical objectives: (i) Accelerating inference speed for real-time fire detection on drones,  (ii) Maximizing memory efficiency to fit large models on the Jetson Nano's limited onboard memory, (iii)  Maintaining high classification accuracy to ensure reliable fire identification, and (iv)  Achieving overall computational efficiency by balancing resource utilization and performance.

To achieve these goals, we present a two-pronged approach:

\begin{enumerate}
    \item Training-time Quantization: We leverage Quantization-Aware Training (QAT) and Automatic Mixed Precision (AMP) to prepare the model for efficient inference. QAT reduces model size and memory footprint by quantifying weights and activations to lower bit widths, while AMP dynamically switches between float and half-precision data representation during training, leading to faster calculations.
    \item Post-training Optimization (PTQ): Beyond QAT and AMP, we explore further optimization techniques for inference on the Jetson Nano.
\end{enumerate}

\begin{itemize}
    \item Hardware-accelerated kernels: Utilizing the Jetson Nano's CUDA cores for parallel execution of computationally intensive operations.
    \item Removing redundant connections and neurons from the model to reduce its size and computational complexity without significant accuracy loss.
\end{itemize}

The significance of these findings lies in the practical implications for fire classification applications on low-power devices. Achieving a balance between computational efficiency and accuracy is crucial for real-world deployment on platforms like the Jetson Nano, where resource constraints are inherent. In the case of classification, the strategic use of FP16 for GPU inference has been shown to offer advantages in terms of memory efficiency and computational speed, while the selection of activation functions like ELU, PRELU, and RELU has been shown to contribute to the non-linear learning capabilities of the model. ELU and PRELU, in particular, can address certain limitations of ReLU by preventing dead neurons and adapting to negative inputs \cite{9461312}. The combination of reduced precision and effective activation functions plays a pivotal role in optimizing model performance in terms of both speed and accuracy. Whereas, performing AMP during the training loop alongside the custom CUDA function block should enable post-training quantization to improve the overall inference without accuracy degradation of the trained model. 

\textbf{Selective quantization.} The final architecture quantization scheme involves the quantization of certain operators to INT8 precision using various calibration methods and granularity, such as per channel or tensor. Residuals, sensitive layers, and non-friendly layers are also quantized to INT8, while other parts of the model remain in FP16 precision \cite{8753553}. This approach offers users significant flexibility in choosing quantization parameters tailored to different network types, enabling the optimization of accuracy and latency simultaneously \cite{9103821}.

\textbf{Memory Efficiency.} An additional focal point of this study is memory efficiency, examined through Active Cache, Active Memory, and Allocator State. These components provide insights into the utilization and allocation of memory during model execution, facilitating a granular analysis of potential bottlenecks and inefficiencies \cite{8753553}. Both qualitative and quantitative observations are taken from the default Pytorch model analyzer.

These methods optimally identify activation functions that strike a balance between computational efficiency and model accuracy \cite{9461312}. The measured latency, throughput variations across different batch sizes, and the impact on model accuracy guide the selection of an optimized configuration for real-time fire classification on the resource-constrained Jetson Nano.

\section{\textbf{Results and Discussions}}
\subsection{Dataset}

This research leverages the FLAME dataset \cite{Flame1data}, a carefully curated collection of images specifically designed for fire classification and segmentation tasks.  The FLAME dataset was collected by two drones during a prescribed fire in an
Arizona pine forest. Choosing the right dataset is important for model performance and generalizability. The dataset comprises 2,003 high-resolution fire images (3480x2160)  captured with a Zenmuse X4S camera, divided into training, validation, and test sets. Each image has a corresponding ground truth mask for fire segmentation. Image augmentation doubles the test data size to 4006 images with masks, all downsampled to 512x512 for training efficiency. FLAME offers several advantages for our research objectives:

\begin{itemize}
    \item Focused on Fire Imagery: Unlike generic image datasets containing a mix of categories, FLAME concentrates solely on fire and non-fire scenarios, ensuring that the model learns specialized features relevant to fire detection.
    \item High-Quality Annotations: FLAME provides accurate and consistent pixel-level annotations for each image, marking the presence or absence of fire at a granular level.
    \item Open-source Availability: FLAME is readily available as an open-source resource, fostering collaboration and reproducibility within the research community.
\end{itemize}

\subsection{Experiment Design}
To incorporate and control activation functions (RELU, ELU, PRELU) in the PyTorch model, the neural network \texttt{nn.ReLU()}, \texttt{nn.ELU()}, and \texttt{nn.PReLU()} functions from the torch library were placed after corresponding layer declarations. During model definition, the desired activation functions are applied to specific layers. The activation parameters and positions based on the network architecture for optimal model behavior are then associated and results are illustrated in Table \ref{tab:activation_metrics} and compared against baselines in Table \ref{tab:activation_baseline}.

\begin{table}[htbp]
    \centering
    \caption{{\small Baseline data for activation functions without CUDA Optimizer {\cite{LI2022145}}.}}
    {\small
    \begin{tabular}{|c|c|c|c|}
        \hline
        Model & MPA (\%) & MIoU (\%) & FPS \\
        \hline
        Deeplabv3+ & 92.09 & 86.75 & 24 \\
        Xceptiondeeplabv3+ & 91.40 & 86.49 & 62 \\
        Fire Segmentation Method & 92.46 & 86.98 & 59 \\
        \hline
    \end{tabular}}
    \label{tab:activation_baseline}
\end{table}

\begin{table}[htbp]
\caption{{\small Activation Function Metrics With Curated CUDA Optimizer on Fire Segmentation Method.}}
\begin{center}
{\small
\begin{tabular}{|c|c|c|c|}
\hline
\textbf{Activation Function} & \textbf{Metric} & \textbf{Validation} & \textbf{Test} \\
\hline
\multirow{4}{*}{ReLU} & Loss & 0.000295 & 0.000289 \\
& MPA (\%) & \textbf{93.4} & \textbf{93.6} \\
& MIoU (\%) & 86 & 85.9 \\
& FPS & \textbf{65.9} & \textbf{66.7} \\
\hline
\multirow{4}{*}{ELU} & Loss & 0.000324 & 0.000324 \\
& MPA (\%) & 93.1 & 93.1 \\
& MIoU (\%) & 84.8 & 84.3 \\
& FPS & 62.4 & 62 \\
\hline
\multirow{4}{*}{PReLu} & Loss & \textbf{0.000289} & \textbf{0.00028} \\
& MPA (\%) & 92.9 & 93 \\
& MIoU (\%) & \textbf{86.3} & \textbf{86} \\
& FPS & 64.3 & 65.6 \\
\hline
\end{tabular}}
\label{tab:activation_metrics}
\end{center}
\end{table}

To leverage CUDA optimization in PyTorch on the Jetson Nano, the CUDA SDK was used to ensure torch and driver compatibility. A custom CUDA optimizer was attached to the model during training for added parallel thread computation. The \texttt{torch.backends.cudnn.benchmark} and \texttt{torch.device('cuda')} functions were used for CuDNN benchmarking and tensor setup respectively. The GPU memory usage was then monitored with adjusted batch sizes on the model architecture for optimal GPU utilization. 

\begin{figure}[htbp]
    \begin{center}
      \includegraphics[scale = .245]{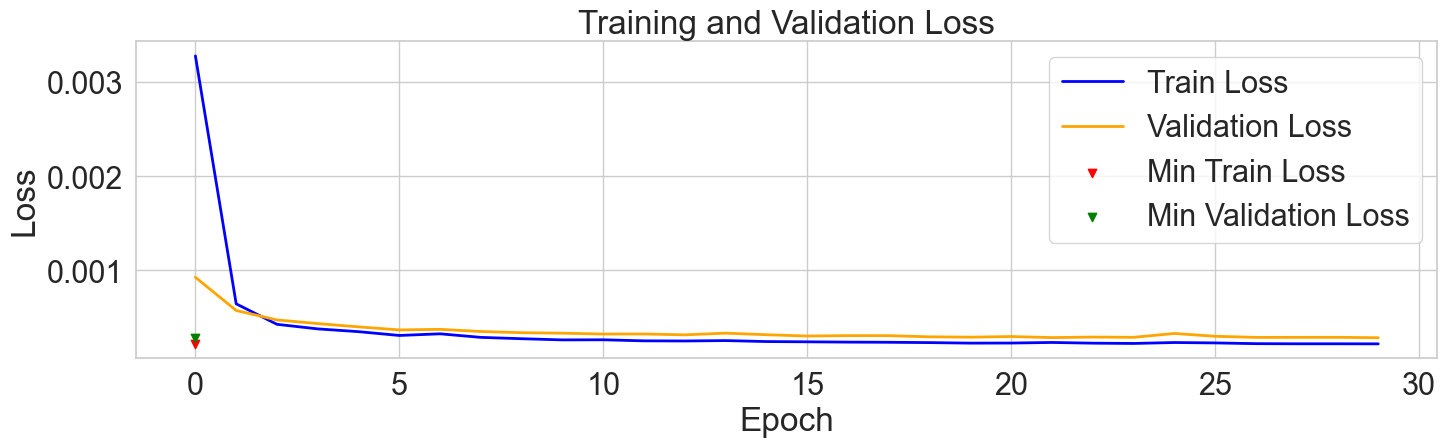}
      \caption{{\small Training and Validation Loss of the ReLu Model. }}
      \label{fig:trainLoss}
  \end{center}
  \vspace{-10pt}
\end{figure}

\label{pseudo}
\textbf{Pseudo-code Procedures for CUDA optimized function block}
{\tiny
\begin{lstlisting}[
  language=Python]
import tensorflow as tf
from compiler.tensorrt import trt_convert as trt

# Define QAT and AMP configuration standard values
qat_calibration_batches = 100
amp_loss_scale = 128

# Training Loop with Quantization Aware Training (QAT) and Automatic Mixed Precision (AMP)
def train_model():
  # Load and preprocess the training data
  train_data, train_labels = preprocess_training_data()
  # Define and compile the model
  model = define_and_compile_model()
  # Apply Quantization Aware Training (QAT)
  qat_model = apply_quantization_training()
  # Apply Automatic Mixed Precision (AMP)
  amp_model = automatic_mixed_precision()
  # Train the model using the mixed-precision optimizer
  train_with_mixed_precision()

# Function to apply Quantization Aware Training(QAT)
def apply_quantization_aware_training():
  # Replace existing nodes with fake quantization of weights
  # Convert activations and compute intermediate tensors
  qat_model = trt.convert()
  return qat_model

# Function to apply Automatic Mixed Precision (AMP)
def apply_automatic_mixed_precision():
  # Reduce memory Requirements and speed up memory operations
  optimizer = CUDA_Optimizer
  amp_optimizer = tf.train.enable_AMP()
  amp_model = tf.keras.models.clone_model()
  amp_model.compile(optimizer= amp_optimizer)
  return amp_model

# Function to train the model using optimizer
def train_with_mixed_precision():
  model.fit()

# Inference Procedure with TF-TRT optimizations
def inference_with_tftrt_optimizations():
  # Load the trained model
  trained_model = load_trained_model()
  # Convert the trained model to TF-TRT optimized model
  trt_optimized_model = trt.convert(trained_model)
  # Run inference using the TF-TRT optimized model
  predictions = trt_optimized_model.predict(data)
  # Process the predictions as needed

# Enable QAT and use AMP
# Call the training loop function
train_model()
# Export the Model
# Load the converted model
# Transfer input data from host to device using cudaMemcpy.
# Call the inference procedure with TF-TRT optimizations
inference_with_tftrt_optimizations()
\end{lstlisting}
}

\begin{figure}[htbp]
    \begin{center}
   \includegraphics[scale = .24]{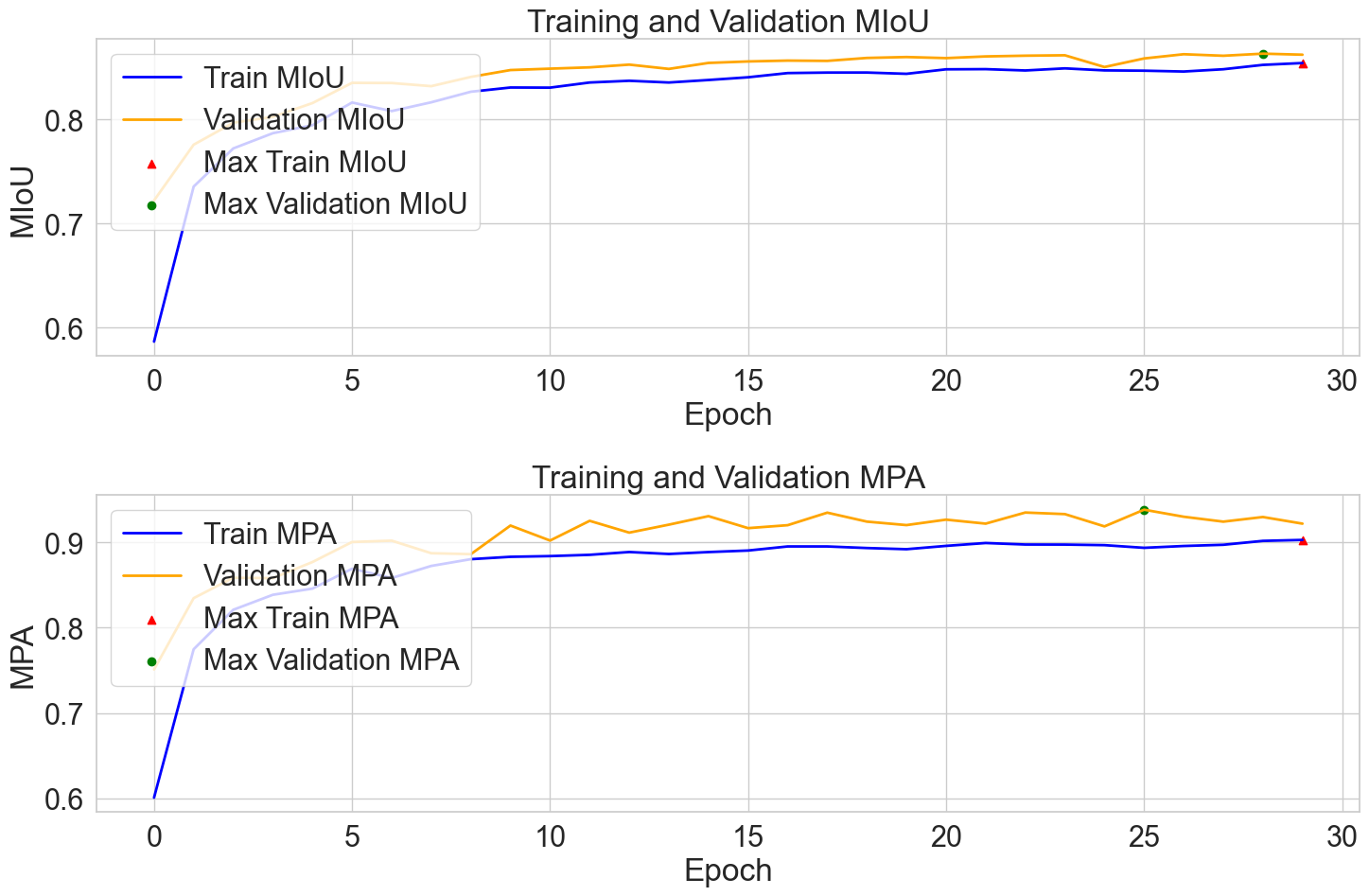}
 %  \subfloat[][Training \& validation MIoU]
 %  {\includegraphics[width=0.49\columnwidth]{Data/Updated_size/MIOU.jpg}}
 % \subfloat[][Training \& validation MPA]
 %  {\includegraphics[width=0.49\columnwidth]{Data/Updated_size/MPA.jpg}}
      \caption{{\small MIoU and MPA for ReLu model training and validation.}}
      \label{fig:trainMIoU}
  \end{center}
  \vspace{-15pt}
\end{figure}

Fig. \ref{fig:trainLoss} shows the training and validation loss of the ReLu model. The loss is seen to remain stable after 10 epochs and relatively lower than other activation loss calculations. Fig. \ref{fig:trainMIoU} depicts ReLu's semantic segmentation model training and validation performance illustrated by the  Mean Intersection over Union (MIoU) and Mean Pixel Accuracy (MPA) metrics.  MIoU measures the intersection of predicted and ground truth regions divided by their union, providing a comprehensive assessment of segmentation accuracy. On the other hand, MPA evaluates the accuracy of individual pixels, representing the ratio of correctly classified pixels to the total number of pixels. Both aggregate values are higher compared to their counterparts. We should note that the results from figures \ref{fig:trainLoss} and \ref{fig:trainMIoU} are of the best quantitative activation function for the relative optimizations. ReLu was shown to outperform the other activation functions on this dataset for both accuracy, mean error, and loss. Algorithm 1 is a code segment of a CUDA kernel for parallel operations.

\begin{figure}[t]
    \begin{center}
      \includegraphics[scale = .1]{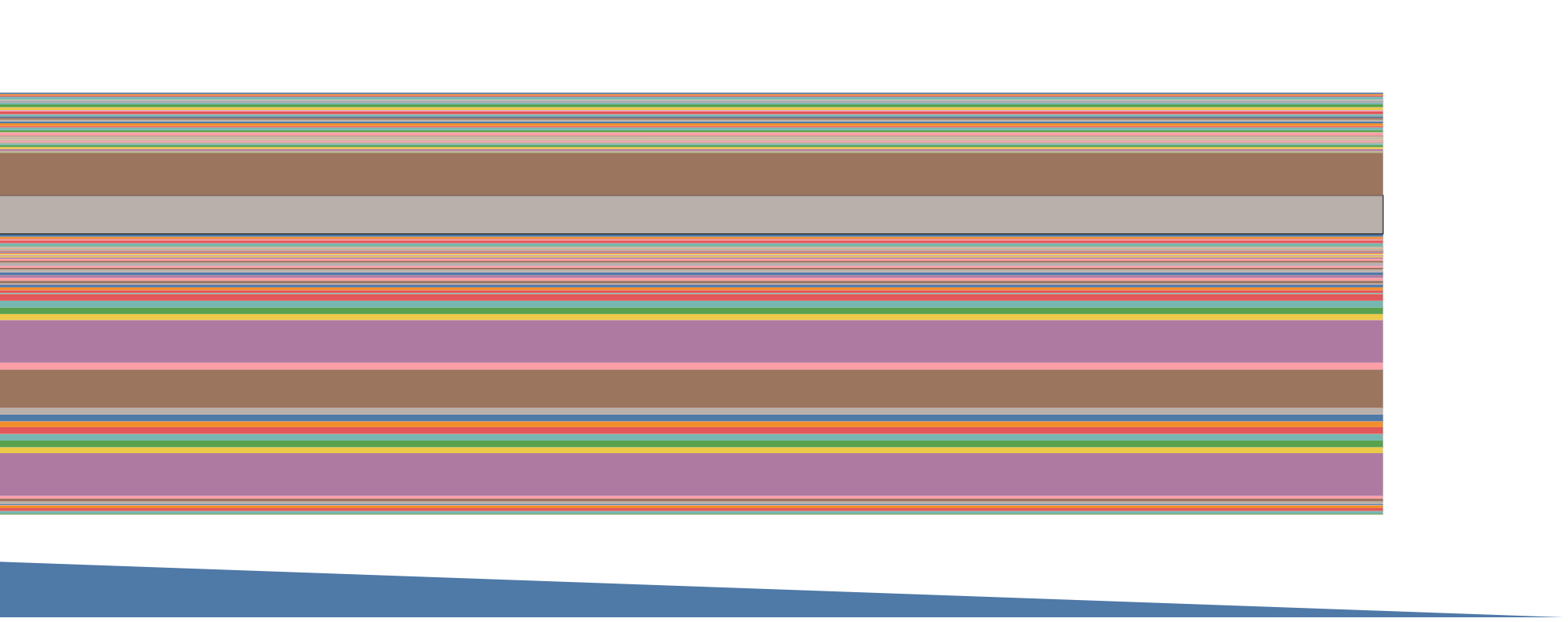}
      \caption{\small{Active memory of the ReLu Model with FP16 quantization where y-axis is time[ms], and x-axis is memory usage[bytes].}}
      \label{fig:active}
  \end{center}
  \vspace{-15pt}
\end{figure}

The active memory, located in Fig. \ref{fig:active}, shows the number of iterations overtime on the x-axis, and the y-axis shows the active memory usage. The total number of iterations was ~90M compared to the non-optimized baseline case of ~175M. This indicates higher memory efficiency in terms of fragmentation and cached memory state, which is important for drone deployments in which have low RAM access. 

\begin{figure}[b]
    \begin{center}
      \includegraphics[scale = .24]{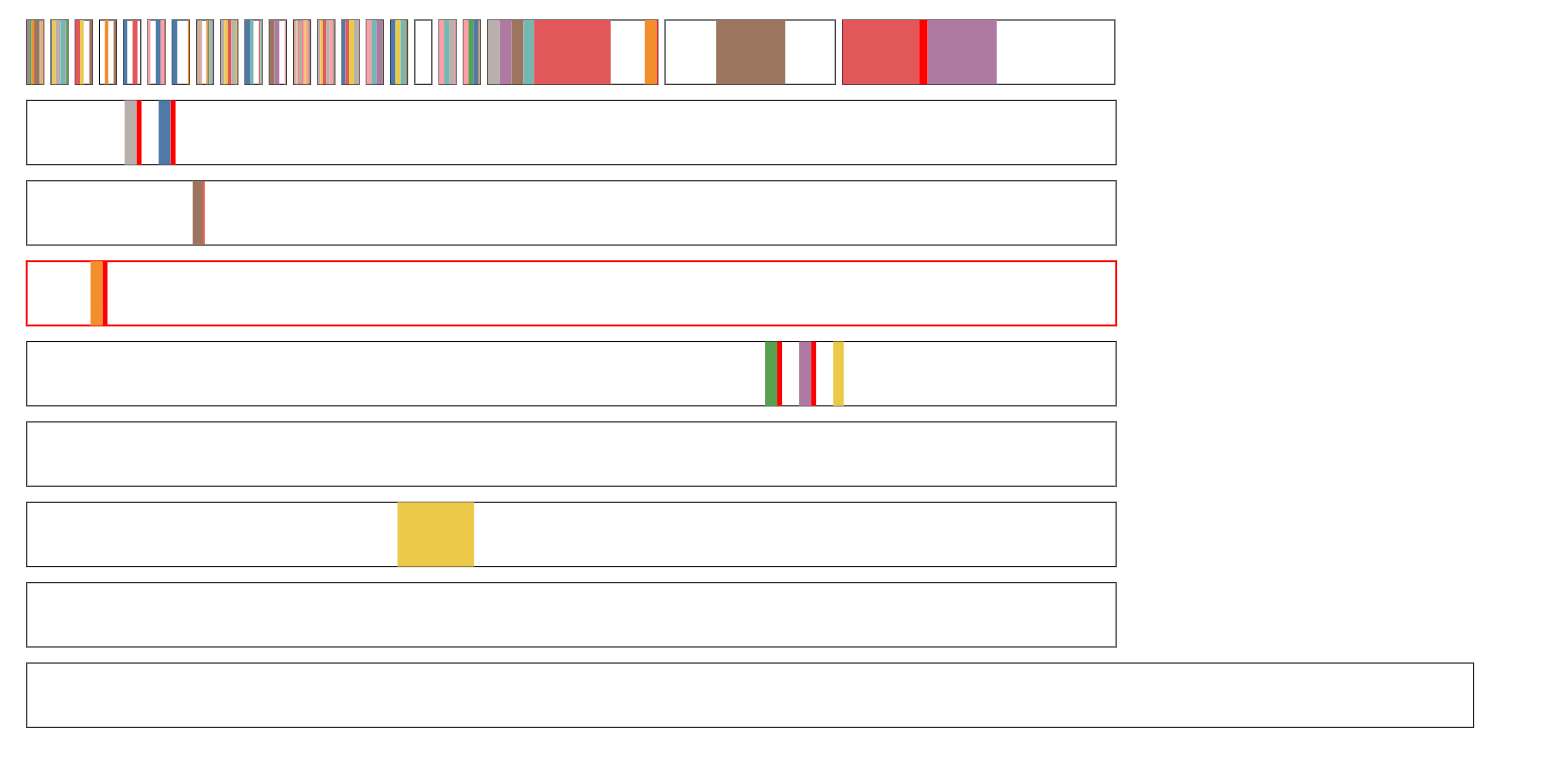}
      \caption{\small{Allocated memory of the ReLu Model with FP16 quantization.}}
       \label{fig:allocated}
  \end{center}
    \vspace{-15pt}
\end{figure}

Fig. \ref{fig:allocated} shows the allocated memory usage (in MB) on the GPU for the best performing activation function, \textbf{ReLu}, and quantization technique \textbf{FP16} on the image classification task using the FLAME dataset. In analyzing fragmentation and cache utilization, spacing, and total colored blocks in the image, ELU and PReLu activation functions were not as efficient as ReLu. ReLu was proven to adequately provide neural network sparsity - memory efficiency in which the model requires less memory to store. The x-axis shows the number of iterations, and the y-axis shows the allocated memory usage. The observation shows fewer allocations compared to both ELU and PReLu activation functions - all of which have a high impact on allocations - and the baseline non-optimized method. Furthermore, there are no relatively excessive allocations.

Quantization involves transforming the deep learning model's parameters to operate at lower precision, reducing model size, and speeding up inference. This optimization is particularly crucial for embedded devices like the NVIDIA Jetson, which have limited computational power. The $precision-mode$ argument available in the TensorFlow-TensorRT SDK (TF-TRT) is used to set the precision mode to FP16. FP16 mode is utilized for Tensor Cores mapped to half-precision hardware instructions. The model was exported with associated sub-graphs by using the TF-TRT SDK. It was then saved via the $saved-model$ format and then converted using the TF-TRT converter engine with batch sizes [2, 32]. 

\begin{figure}[htbp]
    \begin{center}
      \includegraphics[scale = .26]{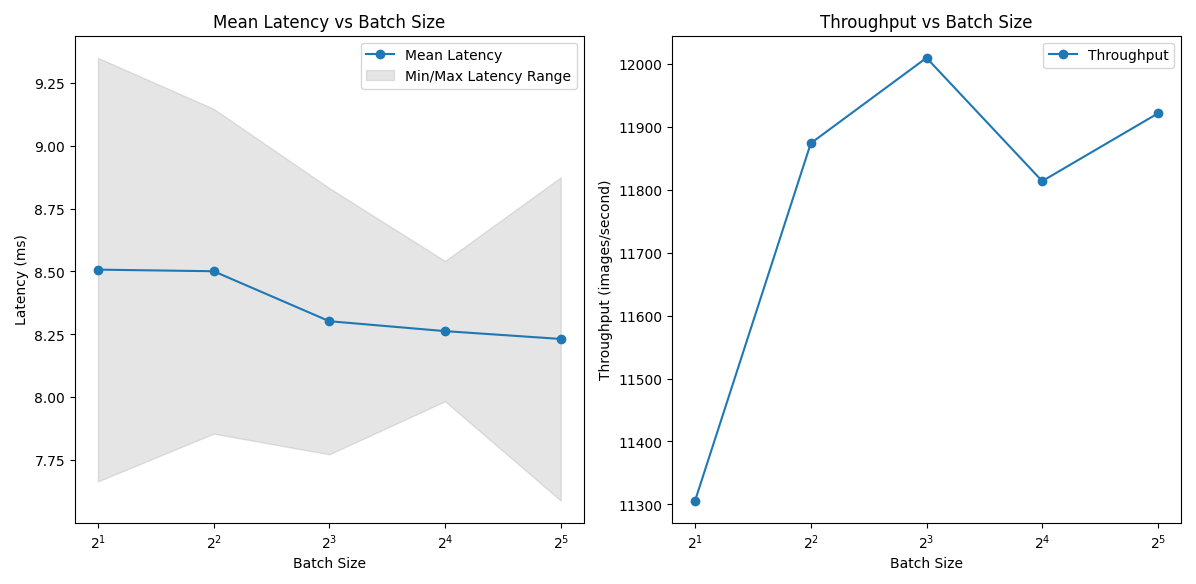}
      \caption{{\small FP16 Quantized Inference of the ReLu Model.}}
      \label{fig:latency}
  \end{center}
  \vspace{-15pt}
\end{figure}

FP16 is seen to improve the overall performance of throughput and latency of the framework. Fig. \ref{fig:latency} shows the quantized inference latency and throughput with Mean Latency: 8.530574083328247 ms, Std Deviation: 1.084523963329141 ms, and Throughput: 12115.00345559036 images/second. The x-axis, batch size, is varied because it affects the training time and generalization accuracy of the model. The y-axis, Latency, and throughput respectively, are the key metrics used when measuring the running model for drone deployment use cases. A batch size of 8 produced the highest throughput with variation in larger batch sizes. Higher batch size trains faster but reduces model performance. Qualitative measurements from the graph show no harsh spiking and solid performance with the provided batch range without substantial accuracy loss.

We should note that {INT8 Quantization} requires supported tensor-core hardware which is not currently available on the P3450. Conversion circuits exist for floating-point and fixed-point accumulation. Thus, FP16 is the primary quantization of interest in this study. Additionally, there is no Post-Training Quantization (PTQ) for the prior fire classification model. Similar models have been shown to have an average throughput of 6400 Images/sec with no quantization \cite{8753553}.

Migration from the Jetson Nano Developer kit into the Jetson Orin or AGX series to utilize NVDLA accelerators and greater capabilities for video encoding with H.265 compression would be of chronological interest. Although the Jetson Nano is more available and cost-efficient, utilizing the NVIDIA Deep Learning Accelerator (DLA), hardware-based acceleration, on supported devices offers significant advantages in terms of power efficiency and robust functionality. DLA's fixed-function accelerator engine accelerates the majority range of neural network layers. The DLA software stack included on supported hardware works in conjunction with TensorRT. TensorRT's higher-level abstractions and combinations alongside DLA should further reduce memory transfers, optimizing performance.

\section{Conclusions}
This paper presents a study on improving early wildfire detection in remote areas using drones with constrained computational and power resources. It develops a real-time image classification and fire segmentation model tailored for efficient functioning on UAVs. The research utilizes hardware acceleration with the Jetson Nano P3450 and investigates the benefits of using TensorRT, a deep-learning inference library. This study systematically explored the impact of activation functions, quantization techniques, and CUDA-accelerated optimizations on deep learning models for image classification, using a UAV-collected forest fire dataset.

Overall, FP16 quantization significantly improved throughput and reduced latency, providing useful insights for optimizing efficiency and accuracy in image fire-segmentation scenarios, with potential applications in drone deployments. The insights gained from this study aim to contribute to the development of efficient and accurate fire segmentation models tailored for edge devices, catering to scenarios where processing capabilities are limited. We believe this work holds promise for advancing real-time, onboard fire detection with drones, empowering quicker responses, via faster inference, and potentially saving lives and ecosystems.

\bibliographystyle{IEEEtran}
\bibliography{ref}

\end{document}